%% file: iclr2024_conference.tex
\documentclass{article} % For LaTeX2e
\usepackage{iclr2024_conference,times}

\input{math_commands.tex}

\usepackage{hyperref}
\usepackage{url}
\usepackage{enumitem}
\usepackage{graphicx}
\usepackage{algorithm}
\usepackage{algpseudocode}
\usepackage{amsmath}
\usepackage{amsfonts}
\iclrfinalcopy
\usepackage{booktabs}
\usepackage{multirow}
\usepackage{colortbl}
\title{Model Tells You Where to Merge: \\ Adaptive KV Cache Merging for LLMs on Long-Context Tasks}

\author{
    Zheng Wang$^{1}$, Boxiao Jin$^{1}$, Zhongzhi Yu$^{1}$, Minjia Zhang$^{2}$\\
    $^1$Georgia Institute of Technology, $^2$University of Illinois Urbana-Champaign\\
    \texttt{\{zwang3478, bjin60, zyu401\}@gatech.edu}, \texttt{\{minjiaz\}@illinois.edu}
}

\algrenewcommand\algorithmiccomment[1]{\hfill \(\triangleright\) #1}
\makeatletter
\newcommand{\algfont}[1]{\renewcommand{\ALG@beginalgorithmic}{\small\selectfont}}
\makeatother

\begin{document}

\maketitle

\begin{abstract}
Large Language Models (LLMs) have attracted remarkable attention due to their unprecedented performance across a wide range of tasks. However, how to efficiently serve LLMs has become a pressing issue because of their huge computational cost in their autoregressive generation process. To mitigate computational costs, LLMs often employ the \emph{KV Cache} technique to improve the generation speed. While improving the computational efficiency, the storage requirements of the KV cache are substantial, particularly in long-context scenarios, leading to significant memory consumption. Existing KV cache eviction methods often degrade the performance of LLMs in long-context scenarios due to the information loss introduced by eviction. In this paper, we propose a novel KV cache merging approach, called \emph{KVMerger}, to achieve adaptive KV cache compression for long-context tasks without significant performance degradation under constrained memory budgets. Our approach is inspired by the intriguing observation that key states exhibit high similarity at the token level within a single sequence. To facilitate merging, we develop an effective yet straightforward merging set identification algorithm to identify suitable KV states for merging. Our merging set identification algorithm stimulates the second observation that KV cache sparsity, from similarity perspective, is independent of the dataset and remains persistent at the model level. Subsequently, we propose a Gaussian kernel weighted merging algorithm to selectively merge all states within each merging set. We conduct extensive experiments to demonstrate the effectiveness of \emph{KVMerger} for long-context tasks under constrained memory budgets, applying it to models including Llama2-7B/13B-chat and Mistral-7B-instruct. Using the LongBench and ZeroScroll benchmarks, we compare our method with other KV cache compression techniques, including H2O and CaM, showing that our method achieves superior performance across tasks with both $50\%$ and $35\%$ KV cache budgets.
\end{abstract}

\section{Introduction}

Large Language Models (LLMs) have demonstrated exceptional performance across a variety of applications, particularly excelling in long-context scenarios that are increasingly relevant in everyday life. Recent state-of-the-art LLMs have been meticulously developed to scale up to handle long-context tasks, such as OpenAI’s ChatGPT \citep{gpt4}, Anthropic's Claude \citep{claude}, Meta’s LLaMA-3 \citep{touvron2023llama} \citep{touvron2023llama2}, Mistral \citep{jiang2023mistral7b},  and Google’s Gemini-pro-1.5 that supports a staggering 1M token context length \citep{geminiteam2024gemini15unlockingmultimodal}. However, as LLMs process larger volumes of data over extended contexts, \emph{KV cache} starts to pose a substantial obstacle to LLM's performance and scalability. KV cache stores the key and value states (KV) derived from the attention calculation of previously processed tokens and reuses those states in the autoregressive generation process. As LLMs continue to grow in size and capabilities, supporting long-context starts to eat up memory. For example, a 175-billion parameter GPT-3 model, with a batch size of 64 and a sequence length of 4,096 tokens (including both prefilled and generated tokens), necessitates approximately 1,208 GB of GPU memory \citep{liu2024minicachekvcachecompression}, which exceeds the memory capacity of most advanced GPUs. Therefore, the need for compressing KV cache while maintaining LLM generation quality, especially for long-context tasks, becomes essential. 

Current efforts for KV cache compression can be broadly categorized into three types: quantization, eviction, and merging, as illustrated in Figure 1. Quantization replaces floating point KV states (e.g., FP16) with low-bit representations to decrease memory usage while striving to maintain the overall performance of LLMs. Recent advancements, such as Coupled Quantization \citep{zhang2024kvcache1bit} and KIVI \citep{https://doi.org/10.13140/rg.2.2.28167.37282}, have demonstrated that KV cache can be quantized to 1-bit or 2-bit precision while preserving performance. In contrast, KV cache eviction methods selectively remove unimportant tokens from the cache based on certain signals from the model, thereby reducing the memory footprint by limiting the number of key and value states in the KV cache \citep{xiao2024efficientstreaminglanguagemodels,liu2023scissorhandsexploitingpersistenceimportance,zhang2023h2oheavyhitteroracleefficient,ge2024modeltellsdiscardadaptive}. 
% One commonly and effectively used signal to identify key and value states to be evicted is the attention score. 
For instance, Scissorhands \citep{liu2023scissorhandsexploitingpersistenceimportance} keeps a fixed KV size budget and replies on the Persistence of Importance hypothesis to evict key and value states for non-important tokens. Similarly, H2O \citep{zhang2023h2oheavyhitteroracleefficient} utilizes aggregated attention scores to determine so called ``heavy hitters", which are a subset of important tokens to keep in the KV cache. While eviction-based methods have demonstrated promising results on short context tasks with simple perplexity metrics, a significant drawback of eviction methods is their potential to accidentally and permanently remove important tokens, leading to context damage and adversely affecting their effectiveness in long-context tasks that heavily rely on context information.  
On a separate line of research, KV cache merging has been proposed as a complementary method of eviction \citep{zhangcam, wan2024d2odynamicdiscriminativeoperations, liu2024minicachekvcachecompression, yu2024effectivelycompresskvheads}.

% \minjia{Are we the first that proposes KV cache merging? If not, seems we need a citation here. Otherwise, people may wonder why we say "KV cache merging has been proposed", and by whom.}

Unlike eviction-based methods, the KV cache merging technique does not strictly remove key and value states. Instead, it involves merging states that are otherwise to be dropped by eviction method into single token state. By amalgamating states rather than outright evicting them, this method ensures that essential information not captured by the attention scores is retained, thereby enhancing the model's ability to maintain performance and accuracy in long-context tasks with compressed KV cache. It is noteworthy that, although token merging is well-established in computer vision (CV) \citep{zeng2022tokensequalhumancentricvisual} \citep{bolya2023tokenmergingvitfaster} \citep{kim2023tokenfusionbridginggap} \citep{zhang2024tinychartefficientchartunderstanding}, the application of key and value states merging in LLMs has not been extensively explored due to several significant challenges. Specifically, \emph{the high dimensionality and sparsity of KV cache make it difficult to accurately identify sets of states that can be merged without losing critical information}. Additionally, \emph{developing appropriate merging algorithm without introducing the loss of essential information in long context presents another major challenge}. Effective merging techniques must strike a delicate balance between reducing memory usage and preserving the semantic integrity of the contexts. 
% \zhongzhi{Shall we use some bold font to highlight the challenges to make it more obvious as these are the key reasons you are doing this work?}

\begin{figure}[t]
    \vspace{-0.5em}
    \centering
    \includegraphics[width=0.85\linewidth]{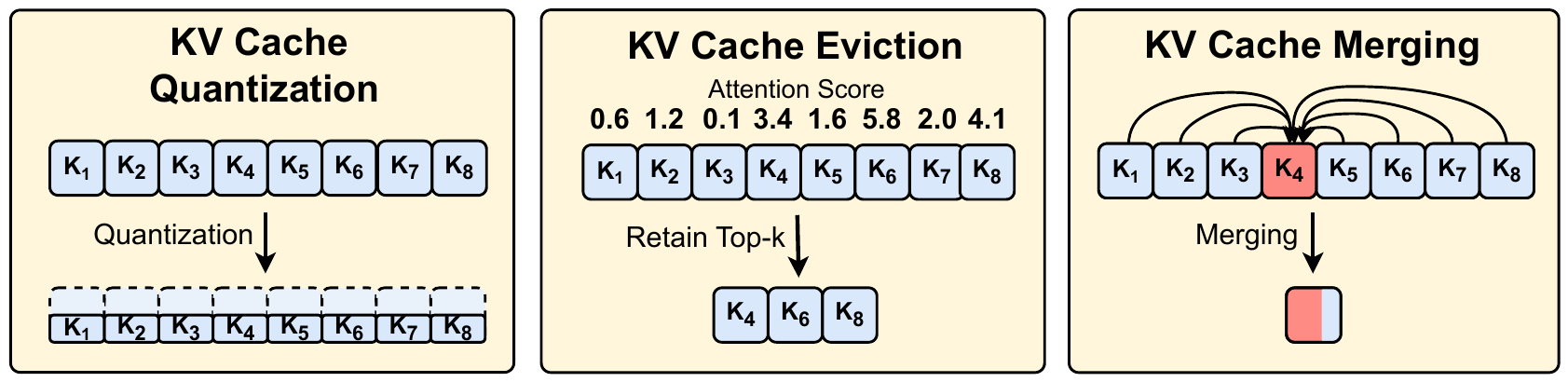}
    \vspace{-0.5em}
    \caption{Three categories of KV cache compression techniques: KV cache quantization (left), KV cache eviction (middle), and KV cache merging (right). For the illustration of KV cache eviction, we use aggregated attention scores as the eviction signal, and k is set to 3; for KV cache merging, we illustrate many-to-one merging. The key state in red represents the state which incorporates the information of other remaining states. Value states are processed in the same way as key states.}
    \label{fig:introduction}
    \vspace{-2em}
\end{figure}

To address the aforementioned challenges associated with KV cache merging, we propose an effective KV cache merging method for accelerating autoregressive LLMs, especially for improving its performance in long-context tasks. We start by introducing an intriguing observation: key states exhibit high cosine similarity at the token level within a single sequence across different attention heads and model layers. 
We investigate the root cause of why such phenomenon appears, and our observation also opens opportunities for effective merging of key and value states based on their cosine similarity.
% \minjia{TODO: The above claim is conditioned on the investigation results of RoPE and token similarity.}
Subsequently, we formulate the KV cache merging as a constrained clustering problem, and we introduce a strong baseline for this problem, where we use an effective merging set identification method for KV cache merging, which results in a layer-wise KV cache compression together with a simple weighted merging algorithm. Based on the proposed merging set identification method, we define KV cache sparsity from the perspective of states similarity. Our finding indicates that KV cache sparsity is independent of the dataset and remains persistent at the model level. Building on top of this, we propose a Gaussian kernel weighted merging algorithm to merge states within each identified merging set. We compare our proposed method with existing KV cache eviction method H2O and value states merging method CaM. The results demonstrate that our method achieves a better performance on these two benchmarks with both $50\%$ and $35\%$ KV cache budgets, surpassing existing KV cache eviction methods. Our contributions can be summarized as follows:
\begin{itemize}[leftmargin=*]
    \item As one of the pioneering researches concerning KV cache merging for LLMs, we developed \emph{KVMerger}, an effective KV cache merging algorithm especially designed for long-context tasks, including merging set identification and Gaussian kernel weighted merging function. 
    \item We introduce an intriguing observation that key states share a high similarity at the token level within a single sequence, as an important complementary to the previous observations concerning high query states similarity \citep{dai2024cormcacheoptimizationrecent} and intra-layer KV cache similarity \citep{liu2024minicachekvcachecompression}. We also investigate the root cause of why such phenomenon appears.
    % \minjia{Shall we also include that we provide analysis on why such similarity pattern exists in LLMs?}
    \item Our proposed \emph{KVMerger} outperforms the previous KV Cache eviction algorithms on long-context tasks across various models under both $50\%$ and $35\%$ KV cache budgets, introducing a great memory reduction compared to full KV cache.
\end{itemize}
% \zhongzhi{Are the contribution bullet points and the above two paragraphs overlapped?}

\section{Related Work and Problem Formulation}
\subsection{KV Cache Quantization}
Quantization methods involve converting high-precision numerical values of key and value states into lower-precision formats, thereby decreasing the storage requirements within the cache~\citep{kvquant,flexgen,llm-qat,q-hitter}. 
% \minjia{There are more quantization work to be cited. I added some, such as kvquant, flexgen, llm-qat, q-hitter. Please double check so we do not miss major ones.}
Due to the presence of outliers in key and value states, recent works such as KIVI \citep{https://doi.org/10.13140/rg.2.2.28167.37282} and Gear \citep{kang2024gearefficientkvcache} employ fine-grained group-wise quantization, which quantize small channel groups within each token. MiKV \citep{yang2024tokenleftbehindreliable} addresses the information loss introduced by KV cache eviction methods by preserving those KVs in lower precision rather than directly dropping them. 
% \minjia{This seems to be a false statement about MiKV. From my understanding, MiKV uses mixed-precision quantization for KV cache and does not evict any tokens. That's why it is called "no token left behind". Please double check.}
ZipCache \citep{he2024zipcacheaccurateefficientkv} proposes an efficient channel-separable quantization scheme, disentangling the channel and token dimensions without excessive memory overhead. Different from quantized KV cache optimizations, this work studies compression of KV cache via token merging, which is complementary to quantization and can lead to better improvements when combined together.

\subsection{KV Cache Eviction}
KV cache eviction methods focus on retaining those important key-value pairs and discard those unimportant ones permanently. One of the common selection policies of key-value pairs is to exploit signals from the attention mechanism of LLMs to select important tokens. For example, H2O \citep{zhang2023h2oheavyhitteroracleefficient}, Scissorhands \citep{liu2023scissorhandsexploitingpersistenceimportance}, and RoCo \citep{ren2024efficacyevictionpolicykeyvalue} compress KV cache by maintaining a small set of KV states whose corresponding tokens are determined by the ranking of attention scores. StreamingLLM \citep{xiao2024efficientstreaminglanguagemodels} finds that keeping the initial tokens, called attention sink,  together with the recent window tokens is pivotal to maintain LLM's performance. More recently, \cite{ge2024modeltellsdiscardadaptive} and \cite{yu2024unveilingharnessinghiddenattention} find that attention sinks also occurs in the middle of the sentences, and \cite{ge2024modeltellsdiscardadaptive} introduces FastGen which can choose the most appropriate compression strategy for each heads with different attention distribution patterns. While demonstrating promising results, existing eviction methods are often evaluated on simple and widely questioned metrics, e.g., perplexity, which may fail to capture LLM's capabilities in understanding long contexts. In contrast, we specifically look into KV compression under more challenging long-context understanding tasks. 

\subsection{KV Cache Merging}
Instead of permanently discarding key and value states, KV cache merging offers a promising direction for KV cache compression while maintaining the performance of LLMs, particularly for long-context tasks such as Retrieval-Augmented Generation (RAG). MiniCache \citep{liu2024minicachekvcachecompression} finds that KV states of some consecutive layers have high similarity and proposes an effective intra-layer KV cache merging and restoration algorithms to reduce memory usage by KV cache. CaM \citep{zhangcam} adaptively merges to-be-evicted value states into the remaining conserved value states, resulting in minimal output perturbation due to the merging operation. Similarly, D2O \cite{wan2024d2odynamicdiscriminativeoperations} selectively merges both value and key states to be evicted with those to be conserved using an Exponential Moving Average (EMA) threshold, and uses weighted merging based on cosine similarity. However, these methods are highly dependent on previous eviction methods, and how to identify effective merging set for KV cache and define effective merging method still remains unclear for KV cache. This paper is the first one to consider KV cache problem independently and propose simple yet effective solutions.

% \minjia{Okay, so there is existing KV merging work. Do we need to include either CaM or D2O as a baseline?}

\subsection{Problem Formulation}
\label{subsec:problem-formulation}
%Formally, we study the performance impact of LLMs after compressing (without fine-tuning) their KV cache. For a pre-trained LLM $f$, we denote its key states and value states as $K\in \mathbb{R}^{n\times d}$ and  $V\in \mathbb{R}^{n\times d}$, respectively. The goal is to develop a KV cache merging algorithm $C$, which consists of two sub-problems: (i) identifying sub KV cache sets with a policy $I$ and (ii) determining the merging policy $M$ for each set that tokens in the same set are merged to a single token by averaging their token key/value states, such that for a task $T$, the resulting compressed LLM $f'$ with merged KV cache leads to no more than $\epsilon$ (compression tolerance threshold) performance drop.  

% Denote key states and value states as $K\in \mathbb{R}^{n\times d}$ and  $V\in \mathbb{R}^{n\times d}$, respectively. The goal is to develop a KV cache merging policy that identifies sub KV cache sets to be merged and determines the merging method for each set, to ensure that the generative results remain similar or comparable to the original ones while reducing KV cache memory.

Formally, we study the performance impact of LLMs after compressing (without fine-tuning) their KV cache. For a decoder only pre-trained LLM $f$, we denote its key states and value states as $\mathcal{K}\in \mathbb{R}^{n\times d}$ and  $\mathcal{V}\in \mathbb{R}^{n\times d}$, respectively. Let $Q_{t}$ denote the query state at time step $t$, and $Q_{t}\in \mathbb{R}^{1\times d}$. Then, the output $\mathcal{O}_{t\ }$ for each attention head at a certain layer of $f$ can be formulated as:
\begin{equation}
\mathcal{O}_{t\ }=\mathcal{A}_{t}\mathcal{V},\  \mathcal{A}_{t}=softmax\left( \frac{Q_{t}\mathcal{K}^{T}}{\sqrt{d_{k}}} \right)
\end{equation}
 \textbf{KV Cache Merging Algorithm.}  Our primary objective is to develop an efficient many-to-one merging algorithm $M$ for KV cache, which should generate merged key states
 % $\mathcal{K}^{*} =\mathcal{K}_{c} \cup M(\mathcal{K}_{m})$
 $M ( \mathcal{K} ) =\bigcup_{i=1}^d F(\mathcal{S}_{ki})$
 % $\mathcal{K}^{*} =\bigcup_{i=1}^d \mathcal{S}_{ki}$ 
 and merged value states 
 % $\mathcal{V} =\bigcup_{i=1}^d \mathcal{S}_{vi}$
 $M ( \mathcal{V} ) =\bigcup_{i=1}^d F(\mathcal{S}_{vi})$, where $\mathcal{S}_{ki}$ and  $\mathcal{S}_{vi}$ represent the sub-merging sets for key states and value states, respectively. $F$ is the merging function which maps the states in each merging set to a single state. Note that  $\mathcal{K} =\bigcup_{i=1}^d \mathcal{S}_{ki}$ and $\mathcal{V} =\bigcup_{i=1}^d \mathcal{S}_{vi}$.
 % where $\mathcal{K}_{c}$ and $\mathcal{K}_{m}$ represent the subsets of key states to be conserved and merged, respectively, and $\mathcal{V}_{c}$ and $\mathcal{V}_{m}$ represent the subsets of value states to be conserved and merged, respectively. Note that $\mathcal{K} =\mathcal{K}_{c} \cup \mathcal{K}_{m}$ and $\mathcal{V} =\mathcal{V}_{c} \cup \mathcal{V}_{m}$. 
 
 \textbf{Definition 2.1} (KV Cache Merging Problem, informal). \emph{Let $\mathcal{O}_{t}$ represent the original output of each attention head at a certain layer of $f$, and let $\mathcal{O}_{t}^{*}$ represent the output after merging. $M$ must satisfy the following optimization criterion:
\begin{equation}
    % M = \arg \min_{M} \frac{|\mathcal{K}^{*}|}{|\mathcal{K}|} = \arg \min_{M} \frac{|M(\mathcal{K})|}{|\mathcal{K}|},
    M = \arg \min_{M} \frac{|M(\mathcal{K})|}{|\mathcal{K}|},
\end{equation}
subject to $|\mathcal{O}_{t}-\mathcal{O}_{t}^{*}| \leq \epsilon$, where $\epsilon$ is an acceptable small positive value, ensuring that the degradation in performance is negligible and within acceptable bounds. $M$ also has the following properties:}
\vspace{-0.5em}
\begin{itemize}[leftmargin=*]
\item $|M\left( \mathcal{K}\right) |\ /\  |\mathcal{K}|\  \leq 1 , |M\left( \mathcal{V} \right)| /\  |\mathcal{V}|\  \leq 1$
\item $|M\left( \mathcal{K}\right) |\ /\  |\mathcal{K}| = |M\left( \mathcal{V} \right)| /\  |\mathcal{V}|\ $ \emph{(make sure key and value states have the same compression ratio)}
\end{itemize} 
\vspace{-0.5em}

In our study, the merging algorithm $M$ consists of two parts: (i) identifying sub KV cache sets with a policy $I$ and (ii) determining the suitable merging function $F$ for each set, where $F$ ensures the states within each set are merged to a single state, such that for a task $T$, the resulting compressed KV cache leads to a performance drop no more than $\epsilon$ (compression tolerance threshold).

\textbf{KV Cache Merging Sets Identification Policy.} We define the identification policy $I$ as: 
\vspace{-0.5em}
\begin{itemize}[leftmargin=*]
\item $|\mathcal{K}| = |\mathcal{K}_{c}| + |\mathcal{K}_{m}|, |\mathcal{V}| = |\mathcal{V}_{c}| + |\mathcal{V}_{m}|$
\item $|\mathcal{K}_{c}| = |\mathcal{V}_{c}| , |\mathcal{K}_{m}| = |\mathcal{V}_{m}|$ \emph{(make sure key states and value states come in pair)}
\end{itemize}
\vspace{-0.5em}
 where $\mathcal{K}_{c}$ and $\mathcal{K}_{m}$ represent the subsets of key states to be conserved and merged, respectively, and $\mathcal{V}_{c}$ and $\mathcal{V}_{m}$ represent the subsets of value states to be conserved and merged, respectively. The above definition is a general formulation. For example, when $|\mathcal{K}_{c}|$ and $|\mathcal{V}_{c}|$ are zero, all key and value states are merged, resulting in a full cache without any states eviction. 

% \minjia{The above seems to be a constraint imposed by the current solution rather than part of the problem formulation. In theory, we can choose to divide the KV cache into arbitrary subsets, e.g., it is possible to merge all tokens into a few super-tokens in certain scenarios. Are these hard constraints really needed for KV cache merging problem formulation?}

\textbf{KV Cache Merging Function.} We define the merging function $F$ such that
% \vspace{-0.5em}
\begin{equation}
% \begin{itemize}[leftmargin=*]
% \item $\left( |M\left( \mathcal{K}_{m} \right) |\  +\  |\mathcal{K}_{c}| \right) \  /\  |\mathcal{K}|\  \leq 1 , \left( |M\left( \mathcal{V}_{m} \right) |\  +\  |\mathcal{V}_{c}| \right) \  /\  |\mathcal{V}|\  \leq 1$
% \item $\left( |M\left( \mathcal{K}_{m} \right) |\  +\  |\mathcal{K}_{c}| \right) \  /\  |\mathcal{K}|\ = \left( |M\left( \mathcal{V}_{m} \right) |\  +\  |\mathcal{V}_{c}| \right) \  /\  |\mathcal{V}|\ $ \emph{(make sure key states and value states have the same compression ratio)}
% \end{itemize} 
F : \{S_i\}_{i=1}^{d} \rightarrow \{s_i^*\}_{i=1}^{d} \quad \text{, where} \quad F(S_i) = s_i^*, \quad i = 1, 2, \ldots, d
\nonumber
\end{equation}
% \vspace{-0.5em}/
where $s_i^*$ is the merged new state for each sub merging set.

% \vspace{-0.5em}
% In this paper, we abstracted KV cache merging sets identification problem as a constrained clustering problem, which is introduced in Section 4.1. We define KV cache merging function as a weighted averaged function in Section 4.2. The proposed $I$ and $F$ follows the above two definitions.
% \zhongzhi{What's the purpose of this sentence if you are going to talk about the observation in the following part?}
\section{Observations}
In this section, we present two key observations illustrating that KV cache sparsity is universal for long-context tasks when viewed from the perspective of state similarity. These observations form the basis for our development of the adaptive KV cache merging algorithm, \emph{KVMerger}.
% \zhongzhi{Try to avoid sentences with multiple clauses. For example, above sentence can be: "In this section, we present two key observations illustrating that KV cache sparsity is universal for long-context tasks when viewed from the perspective of state similarity. These observations form the basis for our development of the adaptive KV cache merging algorithm, KVMerger.", might be more reader-friendly.}
\subsection{KV cache similarity}
Previous literature have analyzed the possibility of reducing KV cache size via exploiting attention sparsity, i.e., identifying important tokens via their attention scores\citep{zhang2023h2oheavyhitteroracleefficient} \citep{liu2023scissorhandsexploitingpersistenceimportance}. However, \emph{attention-score-driven} approaches are biased \citep{he2024zipcacheaccurateefficientkv} because critical tokens often vary a lot across different queries \citep{tang2024questqueryawaresparsityefficient}, where relying on attention-score alone can lead to context damage. Instead of relying on attention scores, we investigate whether merging token states can preserve critical context details. Inspired by \cite{dai2024cormcacheoptimizationrecent},  which reveals the phenomenon that query states share significant similarity at the token level in LLMs, we observe for the first time that \emph{key states also exhibit very high similarity at the token level within single sequence}. We will first demonstrate the generalization of this token level similarity in key states and then analyze the potential reasons behind this intriguing observation.
\begin{figure}[t]
    \vspace{-0.5em}
    \centering
    \includegraphics[width=\linewidth]{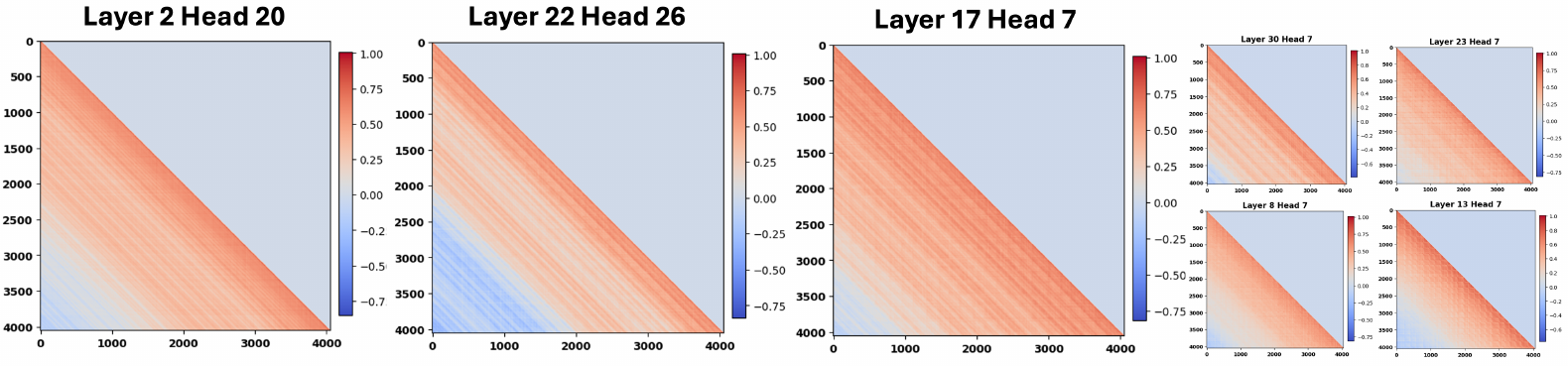}
    \vspace{-1em}
    \caption{Visualization of the cosine similarity map of key states at the token-wise level produced by running the inference process on the Llama2-7b-chat model by randomly sampling data from the SynthWiki dataset. Observations include: (1) Key states share strong similarity within one sequence across different layers and heads; (2) The similarity between key states has the property of locality, i.e., adjacent tokens exhibit higher similarity.}
    \label{fig:similarity_map}
    \vspace{-0.5em}
\end{figure}

\textbf{Observation: key states exhibit high, localized token-level similarity.} We conduct the inference process on the Llama2-7b-chat model by randomly sampling data from the SynthWiki dataset \citep{peysakhovich2023attentionsortingcombatsrecency} with average sequence length being about 4000. Then, we visualize the cosine similarity of key states at the token-wise level within a sequence using the following equation:
\begin{equation}
similarity\left( k_{i},k_{j} \right) =\frac{k_{i}k_{j}^{T}}{||k_{i}||\cdot ||k_{j}||}, \ \ 1\leq i,j\leq T ,
\end{equation}
where $T$ is the total length of the sequence. $k_{i}$ represents the $i$-th key state, and $k_{j}$ represents the $j$-th key state. The visualization results are illustrated in Figure \ref{fig:similarity_map}. We can observe that the similarity maps illustrate a clear oblique color segmentation, and the closer it is to the diagonal, the more intense the color becomes, indicating that key states exhibit a strong localized similarity as query states do \citep{dai2024cormcacheoptimizationrecent}. Specifically, key states share extremely high similarity values with its adjacent tokens, which is greater than $90\%$ for some tokens as Figure 3(a) shows. Moreover, we also observe from Figure 3(a) that the local similarity between one value states and the other consecutive key states shows different fluctuations for different attention heads. We also examine the cosine similarity of value states but do not observe the local similarity property. One interesting question arises: \emph{why do such localized token similarity exhibit in key states, while value states do not?}

\textbf{Analysis.} Recent advancements in large language models (LLMs), including Llama2, Mistral, and Llama3, have showcased significant performance improvements by employing Rotary Position Embedding (RoPE) \citep{su2023roformerenhancedtransformerrotary}. RoPE integrates positional information into token embeddings through a rotational transformation based on positional indices. This process utilizes sinusoidal functions, specifically cosine and sine components, to encode positions. By rotating the embeddings in a multi-dimensional space, RoPE effectively captures the relative positions and order of tokens within a sequence. If we denote two adjacent input tokens as $x_{m} ,\ x_{n} \in \mathbb{R}^{d} $ where $n$ and $m$ are two random integers, then in RoPE, the position information of each token is incorporated via the following equations:
% \minjia{Without loss of generality, should it be $n >= m + 1$ for adjacent input tokens? From Lemma 3.1, it also seems the assumption is |m-n| is small but not necessarily 1.}
\begin{equation}
k_{m,\left[ 2j:2j+1 \right]}\  =\  W_{k}x_{m}e^{im\theta_{j}},\  k_{n,\left[ 2j:2j+1 \right]}=\  W_{k}x_{n}e^{in\theta_{j}},\  \theta_{j} =b^{\frac{-2j}{d}} ,\
\end{equation}
where $W_{k}$ is the matrix for key projection, $b$ is called as the rotary base, which is set to 10000 by default \citep{su2023roformerenhancedtransformerrotary}. 

\textbf{Lemma 3.1} (Informal).
\emph{Consider two vectors $\mathbf{k}_{m}$, $\mathbf{k}_{n} \in \mathbb{R}^{1\times d}$. If their cosine similarity is $1$, then the cosine similarity of any $1 \times 2$ vectors, $\mathbf{k}_{m,j} = \left[ k_{m,2j}, k_{m,2j+1}\right]^T$ and $\mathbf{k}_{n,j} = \left[ k_{n,2j}, k_{n,2j+1}\right]^T$, formed by the $2j$-th and $(2j+1)$-th elements of $\mathbf{k}_{m}$ and $\mathbf{k}_{n}$, $0\leq j \leq (d-1) / 2$, is also equal to $1$.}

\textbf{Lemma 3.2} (Informal). 
\emph{Consider integer $j$ such that $0 \leq j \leq \frac{d-1}{2}$. Define the vectors \( \mathbf{k}_{m,j} \) and \( \mathbf{k}_{n,j} \) as \(\mathbf{k}_{m,j} = \left[k_{m,2j},k_{m,2j+1} \right]^T\) and $\mathbf{k}_{n,j} = \left[ k_{n,2j},k_{n,2j+1} \right]^T$, and define the vectors \( \mathbf{k}_{m,j}^{'} \) and \( \mathbf{k}_{n,j}^{'} \) as $\mathbf{k}_{m,j}^{'} = \mathbf{k}_{m,j} / e^{im\theta_{j}}$ and \(\mathbf{k}_{n,j}^{'} = \mathbf{k}_{n,j} / e^{in\theta_{j}} \). If \(  similarity\left( \mathbf{k}_{m,j}, \mathbf{k}_{n,j} \right) = 1\), we have:
\begin{equation}
    \cos{(m-n)} < \frac{\langle \mathbf{k}_{m,j}^{'}, \mathbf{k}_{n,j}^{'} \rangle}{\|\mathbf{k}_{m,j}^{'}\| \cdot \|\mathbf{k}_{n,j}^{'}\|} \leq 1,
    \nonumber
\end{equation}
where \( \langle \mathbf{k}_{m,j}^{'}, \mathbf{k}_{n,j}^{'} \rangle \) denotes the inner product of \( \mathbf{k}_{m,j}^{'} \) and \( \mathbf{k}_{n,j}^{'} \), and \( \|\mathbf{k}_{m,j}^{'}\| \) and \( \|\mathbf{k}_{n,j}^{'}\| \) denote the norms of \( \mathbf{k}_{m,j}^{'} \) and \( \mathbf{k}_{n,j}^{'} \), respectively.}

% \minjia{Overall good to have a proof here. However, it would be better to add one sentence to indicate the direct takeaway message of Lemma 3.1. I was thinking the Lemma should demonstrate why adjacent tokens x-m and x-n would have their corresponding key states k-m and k-n to have high similarity, e.g., the difference of their cosine distance is within a $\epsilon$, and this $\epsilon$ value depends on the formulation of RoPE. When two tokens have closer position, then the $\epsilon$ value is going to reduce quickly to close to 0 (although need to check if the result is also dependent on x-m and x-n). However, the current Lemma does not seem to directly reflect that, because the conclusion is not about the similarity of key states but how the product of two key states are approximated with their norm. This is an important insight and potentially a good contribution, so please double check.}
The formal and complete proof of the above lemma is shown in appendix \ref{appendixA}. The conclusions of lemmas 3.1 and 3.2 are the necessary conditions of $\textit{similarity}(\mathbf{k}_{m,j}, \mathbf{k}_{n,j})=1$. A cosine similarity of $\mathbf{k}_{m,j}^{'}$ and $\mathbf{k}_{n,j}^{'}$ falling beyond the range $\left( \cos{(m-n)},1\right]$ will result in the failure of $\textit{similarity}(\mathbf{k}_{m,j}, \mathbf{k}_{n,j})=1$.
The analysis above clarifies why value states exhibit low similarity at the token level. Without the RoPE operation, value states are incapable of achieving rotation to comparable angles. Both empirical observations and theoretical analysis indicate that merging highly similar key states is approachable. This scheme is preferable to simply discarding key states, as it helps prevent potential information loss, particularly in long-context scenarios.
% \zhongzhi{I think this part packs too much information. The observation is that the keys are similar while the values are not. Based on my understanding, the major point is the similarity between the keys, and thus removing them doesn't hurt performance. Does it really matter why the values are not similar? The proof about the values may distract the reader from the important information you want to convey. If the proof is an important contribution of the paper, consider first discussing the key takeaways that are valuable for the development of the method at the beginning of this analysis section. After establishing the main points, you can then delve into the details of the proof.}
\begin{figure}[t]
    \vspace{-0.5em}
    \centering
    \includegraphics[width=\linewidth]{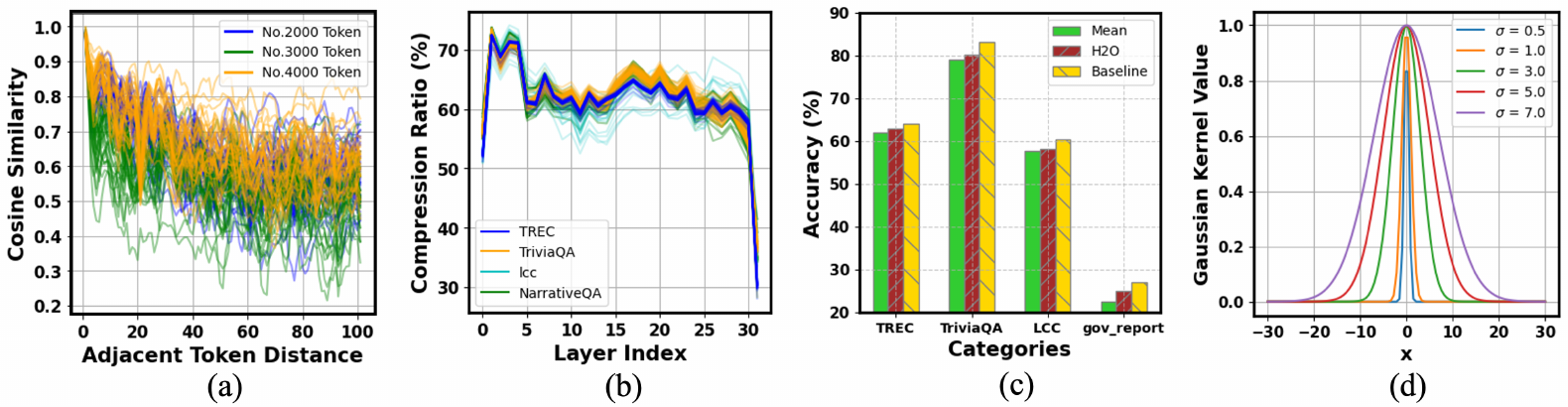}
    \vspace{-1.5em}
    \caption{(a): The cosine similarity changes between the current token and its adjacent tokens across distinct attention
heads and layers. We show the above changes for tokens with indices being 2000, 3000, and 4000.(b) The layer-wise compression ratios obtained by our proposed merging set identification algorithm for different samples and different tasks. (c) The comparison of long-context performance between H2O and average weighted merging with our proposed merging set identification algorithm. (d) The illustration of Gaussian kernel function with different values of $\sigma$.}
    \label{fig:observations}
    \vspace{-0.5em}
\end{figure}
\subsection{Persistent KV cache sparsity}
We have demonstrated that key states within a sequence exhibit significant similarity at the token level in pre-trained LLMs. Based on this, we progressively group consecutive key states of a given key state with similarity values exceeding a certain threshold. By applying this process from the last token to the first token, we obtain a set of groups, each containing consecutive key states with high similarity above the specified threshold. The obtained new key states set is defined as the merging set, meaning that the number of groups in the obtained set equals to the number of key states after merging. The above set identification algorithm is described in detail in Section 4.1.

\textbf{Observation: The KV cache sparsity for different samples are persistent at the model level.}
Figure 3(a) shows that the similarity distributions of different tokens vary across distinct attention heads and layers. The size of each subset of key states is governed by the similarity threshold defined. Lowering the threshold results in the inclusion of a larger number of key states within a single merging set, thereby leading to varied compression ratios across all attention heads and layers. To investigate the actual compression ratio achieved by the previous set identification algorithm, we conduct inference processes on the Llama2-7b-chat model. This involves randomly sampling 200 instances from the subset of LongBench \citep{bai2024longbenchbilingualmultitaskbenchmark} tasks and calculating the average compression ratio for each layer, as shown in Figure 3(b). We observe that the layer-wise compression ratios were highly consistent across different samples from the same task and even across different tasks. This intriguing finding suggests that the \emph{kv cache sparsity, resulting from the high similarity exhibited by key states, is independent of the dataset and remains persistent at the model level}.

\textbf{Insights} The observed static KV cache sparsity suggests that it is possible to determine the layer-wise compression ratios by adjusting the cosine similarity threshold, thereby reducing the KV cache memory consumption. Additionally, Figure 3(b) shows that the first two layers and the last few layers have relatively small compression ratios. This observation aligns with previous research indicating that the attention score distributions are more uniform in the first two layers and last one layer of LLMs \citep{yu2024unveilingharnessinghiddenattention} \citep{wan2024d2odynamicdiscriminativeoperations}, suggesting that most key states are important and should be preserved to avoid introducing significant noise for those layers.
% \minjia{Should we use the term "static KV cache sparsity" or "persistent/universal KV cache sparsity"? My thinking here is that the latter describes the property better because what we are trying to say here is that cache sparsity remain the same across samples/datasets, which is a universal or persistent property rather than a static property. The exact merging sets can still be dynamic for each request, right?}
\begin{figure}[t]
    \vspace{-0.5em}
    \centering
    \includegraphics[width=0.85\linewidth]{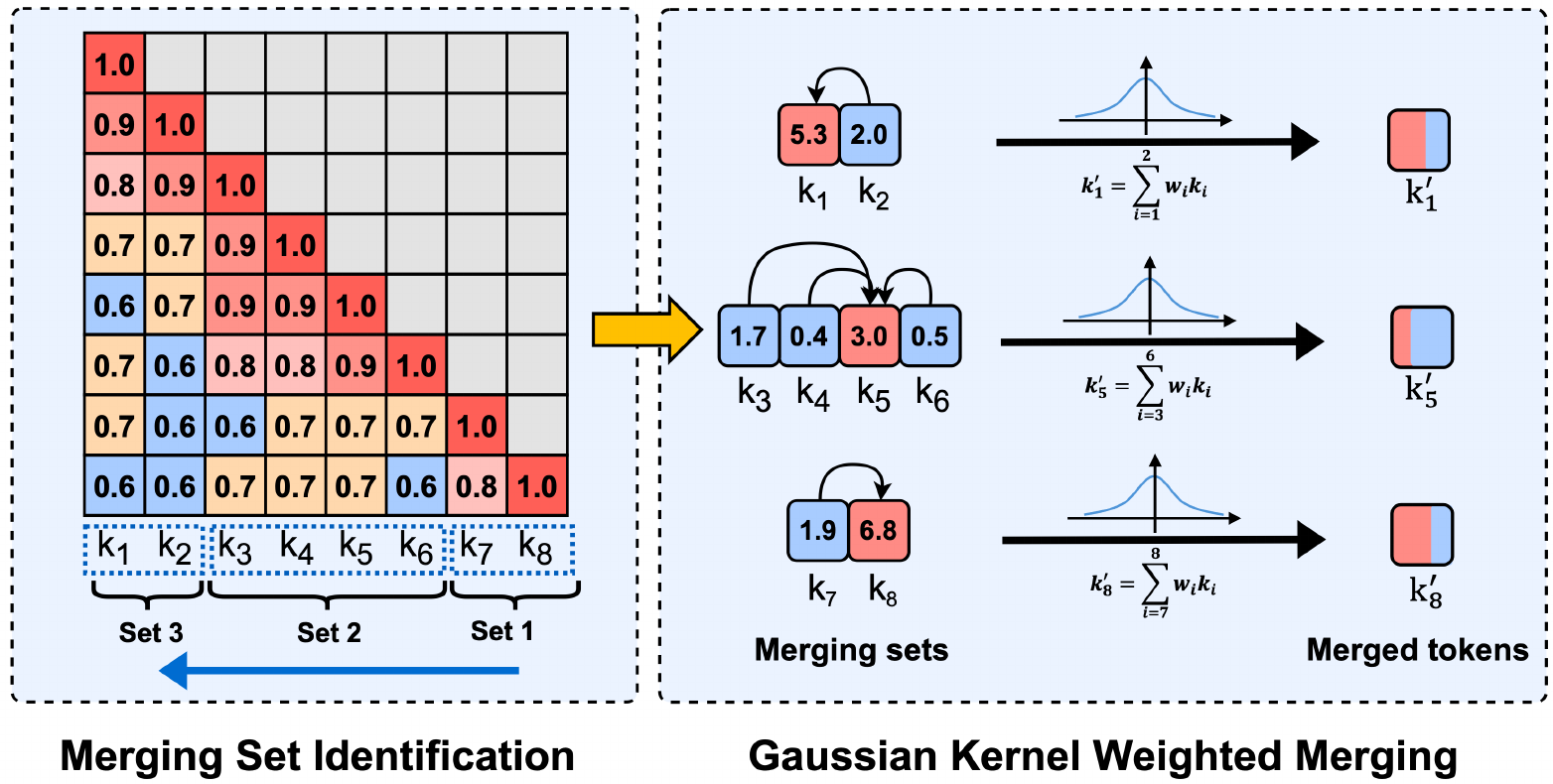}
    \vspace{-1em}
    \caption{The whole framework of \emph{KVMerger} is comprised of two major modules. The first module is to identify the merging set through our proposed algorithm in Section 4.1. Note that those key and value states which are most sensitive to merging are excluded. The toy similarity map is used to illustrate this process in the above Merging Set Identification part, and the threshold for cosine similarity is set to 0.8. The second module is to merge key and value states within each identified merging set via Gaussian kernel weighted merging as described in Section 4.2. For Gaussian kernel weighted merging illustration, the key state in red color represents the pivotal key state, where all the remaining key states should be weighted merged to that one. Note that values on key states in the above graph represent the aggregated attention scores. }
    \label{fig:pipeline}
    \vspace{-1.2em}
\end{figure}
\section{Proposed Adaptive KV Merging Algorithm}
In this section, we propose \emph{KVMerger}, an adaptive KV merging algorithm, for LLMs based on the above observations. The whole pipeline of \emph{KVMerger} is depicted in Figure 4, from which we can see that the whole algorithm contains two major modules: merging set identification and Gaussian kernel weighted merging process. We first introduce the merging set identification algorithm in Section 4.1, which can be viewed as solving a constrained clustering problem. 
% \minjia{This is a bit confusing. We have a problem formulation in Section 2.4, and having a separate problem formulation when describing the solution left people wonder are we solving the same problem or introducing new problems. Suggest to rethink about this part to see it can be merged with Section 2.4 or remove the constraint of clustering problem formulation to avoid confusion.}
We propose a transformation of Agglomerative Hierarchical Clustering (AHC) algorithm to solve this. In Section 4.2, we delineate our proposed Gaussian kernel weighted merging algorithm, which is a many-to-one states merging method without introducing significant information loss. 

\subsection{Greedy Policy for Merging Set Identification}
One way to solve the merging set identification problem described in Section~\ref{subsec:problem-formulation} is to view it as a variant of clustering problem, which we define below:

% \textbf{Definition 4.1} (Clustering Problem, formal). \emph{Given a set of data points \(\mathcal{D} = \{x_1, x_2, \ldots, x_n\}\), where each \(x_i \in \mathbb{R}^d\), and a similarity (or distance) function \(\delta: \mathcal{D} \times \mathcal{D} \rightarrow \mathbb{R}\), partition \(\mathcal{D}\) into \(k\) clusters \(\{C_1, C_2, \ldots, C_k\}\) such that the intra-cluster similarity is maximized and the inter-cluster similarity is minimized.}
% \begin{itemize}[leftmargin=*]
%     \item \emph{Each cluster \(C_i\) should satisfy: $C_i \cap C_j = \emptyset \quad \text{for} \quad i \neq j, \quad \text{and} \quad \bigcup_{i=1}^k C_i = \mathcal{D}$}
%     \item \emph{The objective function to be maximized can be expressed as:}
    
%     $\ \ \ \ \ \ \  \ \ \ \ \  \ \ \ \ \ \max_{C_1, C_2, \ldots, C_k} \left( \sum_{i=1}^k \sum_{x, y \in C_i} \delta(x, y) - \sum_{i \neq j} \sum_{x \in C_i, y \in C_j} \delta(x, y) \right)$
% \end{itemize}

\textbf{Definition 4.1} (Constrained Clustering Problem for KV Cache Merging, formal). \emph{Given the original set of key states to be merged \(\mathcal{K}_{m} = \{k_1, k_2, \ldots, k_n\}\) from a certain attention head at a certain layer of $f$, where each \(k_n \in \mathbb{R}^{1\times d}\), and a similarity function \(\delta: \mathcal{K}_{m} \times \mathcal{K}_{m} \rightarrow \mathbb{R}^{n\times n}\), partition \(\mathcal{K}_{m}\) into \(d\) merging sets \(\{\mathcal{S}_1, \mathcal{S}_2, \ldots, \mathcal{S}_d\}\) such that the intra-cluster similarity is maximized and the inter-cluster similarity is minimized.}
\vspace{-0.8em}
\begin{itemize}[leftmargin=*]
    \item \emph{Each merging set \(\mathcal{S}_i\) should satisfy: $\mathcal{S}_i \cap S_j = \emptyset \quad \text{for} \quad i \neq j, \quad \text{and} \quad \bigcup_{i=1}^d \mathcal{S}_i = \mathcal{K}_{m}$;}
    \item \emph{$\forall \mathcal{S}_{i}$, $\exists j$ such that  $\mathcal{S}_{i} = \{ k_{j}, k_{j+1}, \cdots, k_{j + |\mathcal{S}_i| - 1} \}$;}
    % \item \emph{$\forall i, \ \exists j_i \ \text{such that states' indices of } \ \mathcal{S}_i = \{ j_i, j_i + 1, \ldots, j_i + |\mathcal{S}_i| - 1 \}$, where $ j_{i}$ is the starting index of \(\mathcal{S}_i\);}
    \item \emph{The objective function to be maximized can be expressed as:
    \begin{equation}
        \max_{\mathcal{S}_1, \mathcal{S}_2, \ldots, \mathcal{S}_d} \left( \sum_{i=1}^d \sum_{k_{a}, k_{b} \in \mathcal{S}_i} \delta(k_{a}, k_{b}) - \sum_{(i,j),i \neq j}^d \sum_{k_{a} \in \mathcal{S}_i, k_{b} \in \mathcal{S}_j} \delta(k_{a}, k_{b}) \right)
        \nonumber
    \end{equation}}
\end{itemize}
\vspace{-0.5em}
 % \minjia{The above definition seems to be a generic definition of clustering, which probably is too general for this paper. Suggest to directly define the constrained clustering problem for the merging set identification problem by incorporating the constrains described below. The name of the definition can be "Constrained Clustering for KV Cache Merging". After that, the text can directly introduce the AHC algorithm.}
 % \minjia{This formulation of constrained clustering problem makes me also think the problem formulation for generic KV cache merging in Section 2.4 can be more general, e.g., without adding strong constraints. This way, while we view it as a constrained clustering optimization problem in this paper, other people may inherit the problem definition in Section 2.4 but with different solutions for KV cache merging.}
 The similarity function, $\delta$, we used here is cosine similarity based on the observation in Section 3.1. In order to conserve the locality similarity property of key states, the merging set identification problem is a constrained clustering problem, meaning that all elements in one cluster should be consecutive in sequence, and we do not merge states with high similarity but far away from each other. Then, we propose a variant of Agglomerative Hierarchical Clustering (AHC) algorithm to find all merging sets shown as Algorithm 1. 
 
\begin{minipage}[t]{0.493\textwidth}
\vspace{-1.5em}
\begin{algorithm}[H]
\caption{Merging Set Identification}
\label{alg:demo1}
\begin{algorithmic}[1]
\Procedure{AHC}{$\mathcal{K}_{m} = \{k_1, \ldots, k_{T}\}$, $\delta$, $\epsilon$}
    \State Start with T clusters with one key state 
     % \Statex \hspace{1.9em}\(C_i = \{k_i\}\)
    \State Compute the similarity matrix \(\Delta\) where 
    \Statex \hspace{1.9em}\(\Delta(i, j) = \delta(k_i, k_j)\)
    \For{$i = T \to 1$}
        \State Group \(\{k_i\} \cup \{k_j \mid \delta(k_i, k_j) > \epsilon \},\) 
        \Statex \hspace{2.8em} s.t. \( \ || i - j || = 1\) 
    \State  i = j
    \EndFor
    \State \Return The merging sets
\EndProcedure
\end{algorithmic}
\end{algorithm}
\end{minipage}
\hfill
\begin{minipage}[t]{0.493\textwidth}
\vspace{-1.5em}
\begin{algorithm}[H]
\caption{Merging Policy}
\label{alg:demo2}
\begin{algorithmic}[1]
\Procedure{Merge}{$\mathcal{S}_{k} = \{k_1, \ldots, k_{n}\}, A$}
    \State Compute aggregated attn score for each 
    \Statex \hspace{1.9em}state: $\mathbf{a}_{i}=\sum_{i=1}^{|S_{k}|} A\left[ i,: \right]$
    \State Find pivotal state: ${p} = \underset{i \in S_{k}}{\text{argmax}}\, (\mathbf{a}_{i})$
    \For{$j = 1 \to n$}
        \State $\mathbf{g}_{pj} = \mathcal{G}(\mathbf{k}_{p}, \mathbf{k}_{j})$
    \EndFor
    \State $k_{M} = \sum_{}\mathbf{w}_{j}k_j, \ \  \mathbf{w}_{j}=\mathbf{g}_{pj} / \sum_{}\mathbf{g}_{pj}$
    % \State$k_{M} = \mathbf{w}_{1}k_1 + \ldots + \mathbf{w}_{m}k_m$
    \State \Return $k_{M}$
\EndProcedure
\end{algorithmic}
\end{algorithm}
\end{minipage}
\\

\emph{KVMerger} also retains the KV states whose corresponding aggregated attention scores fall within the top-k range, including both attention sinks and heavy-hitters, which represent the most important and frequently accessed elements by LLMs. We assume that those key and value states are quite sensitive to merging and cannot participant in merging process to avoid information damage. 

\subsection{Gaussian Kernel Weighted Merging}
\textbf{Definition 4.2} (Weighted KV cache Merging, formal). \emph{Given identified merging sets of key states and value states as \(\mathcal{S}_{k} = \{k_i, k_{i+1}, \ldots,k_{p}, \ldots k_{i+n}\}\) and \(\mathcal{S}_{v} = \{v_i, v_{i+1}, \ldots,v_{p}, \ldots v_{i+n}\}\) , where $k_{p}$ and $v_{p}$ denote the pivotal key state and pivotal value state, respectively. Then, the weighted merging key states and value states can be defined as:
\begin{equation}
k_{p} = w_{p}k_{p} + \sum_{k_{i} \in \mathcal{S}_{k}, i \neq p} w_{i} k_{i}, \quad
v_{p} = w_{p}v_{p} + \sum_{v_{i} \in \mathcal{S}_{v}, i \neq p} w_{i} v_{i}
\end{equation}
where $w_{p}$ and $w_{i} $ denote the weight assigned to the pivotal state and the remaining states in the merging set. }

We define the weighted merging function for KV cache merging in Definition 4.2, which follows the many-to-one merging definitions from \cite{wan2024d2odynamicdiscriminativeoperations}. Note that the merging function 
% \minjia{We need to unify the term a bit. Right now, we are using merging policy (in Section 2.4), merging algorithm (overall approach), and merging methodology (here) in different places. Let's use merging algorithm $C$ to refer to the overall approach, merging set identification ($I$) to indicate the set selection process/algorithm, and merging policy/function ($M$) to refer to the function for merging states?}
is a critical determinant of performance in many-to-one merging scenario. Two principal design factors directly influence merging efficacy. The first factor is the selection of the pivotal state, to which all other states are merged. 
% \minjia{TODO: This is the first time "pivotal tokens" and "assignment of weights" appear. It would be better if we add the description of assignment of weights in problem formulation in Section 2.4 when describing the merging policy $M$. The challenge is how to assign the weights.}
The second factor involves the assignment of weights to each state, with the pivot
state having the largest weight to preserve the information.
% Specifically, the pivotal token is typically assigned the largest weight within the merging set. If the pivotal token is not the most informative one, a substantial amount of weight will be allocated to it, leading to potential information loss and bias. \zhongzhi{How about just mentioning that pivot needs to be the most informative one, instead of saying the counterpart, which is a bit confusing. An example flow is: The first factor is the selection of the most informative pivot token, to which all other tokens are merged. The second factor involves the assignment of weights to each token, with the pivot token having the largest weight to preserve the information.}

% \minjia{The above description of weighted merging seems to belong to the problem formulation in Section 2.4. Another option is to keep the definition of 2.4 generic, e.g., without adding the constraints, and then describe the weighted merging mechanism here in the design section. }
We start from the most intuitive merging function via average weighted for each merging set. We evaluate the average weighted merging function on four tasks from LongBench: TREC, NarrativeQA, TriviaQA, and LCC. As highlighted in previous research \citep{xiao2024efficientstreaminglanguagemodels} \citep{zhang2023h2oheavyhitteroracleefficient} \citep{wan2024d2odynamicdiscriminativeoperations}, recent tokens play a crucial role in performance. Therefore, we exclude the most recent tokens from merging to maintain an average compression ratio of $50\%$, as discussed in Section 3.2. For simplicity, we select the pivotal token as the key state with the maximum index within each merging set. Additionally, we compare the average weighted merging function with the H2O algorithm to gain an initial perspective on the potential and performance differences between the KV cache merging scheme and the eviction scheme. The evaluation results are shown in Figure 3(c). The results demonstrate that the average weighted merging scheme provides a robust baseline, affirming the efficacy of the current method for identifying merging sets. However, the average weighted merging function performs worse compared to H2O, suggesting that the merging process may introduce noise, leading to information distortion.
% \zhongzhi{Looks like this paragraph is about a baseline method for observation part. If this is the case, why not describe the method in sec 3.2 and only the key insight and limitation here to motivate the real method in the following paragraph? }

% \begin{figure}[t]
%     \vspace{-0.5em}
%     \centering
%     \includegraphics[width=0.85\linewidth]{framework.pdf}
%     \vspace{-1em}
%     \caption{\zhongzhi{Shall we add some arrows to show the flow in this figure?}The whole framework of \emph{KVMerger} is comprised of three major modules. The first module is to identify key and value states are most sensitive to merging and exclude them from merging. We use top-k heavy-hitters to identify those states. The second module is to identify the merging set through our proposed algorithm in Section 4.1. The toy similarity map is used to illustrate this process in the above Merging Set Identification part. The third module is to merge key and value states within each identified merging set via Gaussian kernel weighted merging as described in Section 4.2. For Gaussian kernel weighted merging illustration, the key state in red color represents the pivotal key state, where all the remaining key states should be weighted merged to that one. }
%     \label{fig:pipeline}
%     \vspace{-1.2em}
% \end{figure}

\textbf{Gaussian Kernel Weights} To eliminate the noise introduced by less informative key states via average weighted merging, we introduce Gaussian kernel weighted merging, which is expressed as:
\begin{equation}
\mathbf{g}_{pi}=\mathcal{G}(\mathbf{k}_{p}, \mathbf{k}_{i})=exp\left( -\frac{||\mathbf{k}_{p}-\mathbf{k}_{i}||^{2}}{2\sigma^{2}} \right) ,\  \  \ \ \  \sigma =\frac{\sum_{0}^{|\mathbf{S}_{k}|} \mathbf{g}_{pi}}{\sqrt{2}|\mathbf{S}_{k}|}.
\end{equation}

% \minjia{TODO: Need one sentence to explain the intuition of why Gaussian kernel weighted merging is a good solution instead of just being a solution to address the limitation of naive weighted sum merging.}
Gaussian kernel is able to assign greater weight to elements that are nearer to the pivotal state. This local weighting characteristic ensures that the merged result is significantly shaped by nearby states, maintaining local structure and minimizing the impact of distant, possibly noisy states. Then, the merging weights for key states and value states can be formalized as:
\begin{equation}
\mathbf{w}_{i}=\frac{\mathbf{g}_{pi}}{\sum_{j=0}^{|S_{k}|} \mathbf{g}_{pj}}, \ \ \ \ \  \mathbf{w}_{p}=\frac{1}{\sum_{j=0}^{|S_{k}|} \mathbf{g}_{pj}}. 
\end{equation}
For merging value states, $\mathbf{w}_{i}$ and $\mathbf{w}_{p}$ should also multiple $|S_{v}|$. This adjustment is necessitated by the need to accurately reflect the number of value states during the merging process, ensuring that the merged result accurately represents the contribution of each value state. As demonstrated in Definition 4.2, the weight assigned to each $\mathbf{k}_{i}$ and $\mathbf{v}_{i}$ is directly governed by the squared $l_{2}$ norm between the pivotal token and the remaining tokens. This indicates that if $\mathbf{k}_{i}$ is close to $\mathbf{k}_{p}$ in the Euclidean space, more weight will be assigned to $\mathbf{k}_{i}$ as Figure 3(d) illustrates. Therefore, the value of $\sigma$ is crucial as it directly determine the weights. Specifically, if $\sigma$ approaches 0 and $||\mathbf{k}_{p}-\mathbf{k}_{i}||^{2}$ is significantly different from 0, the weight assigned to $\mathbf{k}_{i}$ tends towards 0. Consequently, no additional information is preserved except for the pivotal token. We empirically define $\sigma$ as the mean value of $\mathbf{g}_{pi}$ for all tokens within each merging set to avoid such situation. 

\textbf{Selection for Pivotal State} As previously discussed, the selection for pivotal state within each merging set is crucial for performance. Here we follow previous token eviction methods that using aggregated attention score to select pivotal token as it indicates the importance of tokens, which can be expressed as:
\vspace{-1em}
\begin{equation}
\mathbf{k}_{p} = \underset{i \in S_{k}}{\text{argmax}} \, (\mathbf{a}_{i}),   \ \ \ \ \ \mathbf{a}_{i}=\sum_{i=0}^{|S_{k}|} A\left[ i,: \right]
\end{equation}
% \vspace{-0.5em}
Note that the index of pivotal token for value states within each merging set is the same as key states. The complete merging policy is described as Algorithm 2. 

% \subsection{Overview of the Adaptive KV Cache Merging}
% \zhongzhi{Overivew is usually at the beginning of the method section. Do you have specific reason why placing it here? }
% The proposed KV cache merging algorithm is depicted in Figure 4. Beyond the merging set identification and the Gaussian kernel weighted merging function, \emph{KVMerger} also retains the KV states whose corresponding aggregated attention scores fall within the top-k range, including both attention sinks and heavy-hitters, which represent the most important and frequently accessed elements by LLMs. We assume that those key and value states are quite sensitive to merging and cannot participant in merging process to avoid information damage. By doing so, the algorithm ensures that the most critical and frequently referenced data points are readily accessible, thereby enhancing overall system performance and responsiveness with the joint efforts of other merged states.

\section{Experiment}
\subsection{Experimental Settings}
\textbf{Models and Tasks} We evaluate \emph{KVMerger} using three models: Llama2-7B/13B-chat \citep{touvron2023llama2} and Mistral-7B-Instruct-v1.0\citep{jiang2023mistral7b}. Our evaluation focuses primarily on instruction-tuned models, as these are meticulously optimized for dialogue use cases and question-answering scenarios. The above three models are evaluated on two commonly used benchamrks for long-context scenario, that is, LongBench \citep{bai2024longbenchbilingualmultitaskbenchmark} and ZeroScrolls \citep{shaham2023zeroscrollszeroshotbenchmarklong}. Both LongBench and ZeroScrolls include a variety of scenarios such as multi-document question answering, summarization, and dialogue generation, providing a comprehensive assessment of a model's ability to handle long sequences of text while maintaining coherence and accuracy. Specifically, we use nine datasets in LongBench: 2WikiMQA, gov$\_$report, NarrativeQA, passage$\_$retrieval$\_$en, MultifieldQA$\_$en, TREC, multi$\_$news, TriviaQA, qasper.
We use seven datasets in ZeroScrolls: gov$\_$report, SummScreenFD, QMSum, SQuALITY, Qasper, NarrativeQA, BookSumSort. 
Additionally, we also individually test our methods on RAG tasks with the Needle-in-a-Haystack test \citep{guerreiro2023lookingneedlehaystackcomprehensive}. 
The performance of our method for LLMs on all the above tasks are also compared with existing eviction method H2O and merging method CaM. 

\textbf{Implementation details} We test \emph{KVMerger} in two compression scenarios. The first one is $50\%$ KV cache budget, where the proportion of recent tokens to be reserved is set to $0.17\%$, and the proportion of states not selected for the merging process in terms of aggregated attention scores is set to $0.12\%$. The remaining key states and value states participate in the merging process. The second compression scenario is $35\%$ KV cache budget, where the proportion of recent tokens is set to $0.08\%$, and the proportion of states not selected for the merging process is set to $0.02\%$. The cosine similarity threshold for both two scenarios is set to $0.75$. We conducted our experiments on a cluster with A100 40GB GPUs (4XA100 per node, 256GB DRAM, 15TB storage, 200Gbps interconnect), and a cluster with A100 80GB GPUs (2xA100 per node, 256GB DRAM, 100Gb Ethernet interconnect). The evaluation process for LongBench and ZeroScrolls follows \cite{longbench} and \cite{zero_scrolls}. The implementation of Needle-in-a-Haystack test follows \cite{kamradt_needle_haystack}.

\begin{table}[t]
\vspace{-1em}
\caption{\emph{KVMerger} for Llama2-7B/13B-chat and Mistral-7B-Instruct on \textbf{LongBench} datasets.}\label{tab:LongBench}
\large
\resizebox{1\textwidth}{!}{
\begin{tabular}{@{}ccccccccccccc@{}}
\toprule
\hline
\textbf{models}                            & \textbf{budget}                  & \textbf{method}                             & \textbf{2wikimqa}                      & \textbf{gov\_report}                   & \textbf{narrativeqa}                   & \textbf{pr\_en}                        & \textbf{multifieldqa\_en}              & \textbf{trec}                          & \textbf{multi\_news}                   & \textbf{triviaqa}                      & \textbf{qasper}                        & \textbf{avg.}                       \\ \midrule \hline
                                  & \cellcolor[HTML]{DAE8FC}100\% & \cellcolor[HTML]{DAE8FC}Full Cache & \cellcolor[HTML]{DAE8FC}31.45 & \cellcolor[HTML]{DAE8FC}26.99 & \cellcolor[HTML]{DAE8FC}18.74 & \cellcolor[HTML]{DAE8FC}8.00  & \cellcolor[HTML]{DAE8FC}36.60 & \cellcolor[HTML]{DAE8FC}64.00 & \cellcolor[HTML]{DAE8FC}26.26 & \cellcolor[HTML]{DAE8FC}83.09 & \cellcolor[HTML]{DAE8FC}21.83 & \cellcolor[HTML]{DAE8FC}35.22 \\ \cmidrule(l){2-13} 
                                  &                               & H2O                                & 29.96                         & 24.86                         & 17.48                         & 7.00                          & 33.58                         & 63.50                         & 26.00                         & 82.51                         & \textbf{21.04}                & 34.00                         \\ \cmidrule(l){3-13} 
                                  &                               & CaM                                & 30.69                         & 24.46                         & 17.08                         & 6.50                          & 33.98                         & 63.50                         & 24.66                         & 82.17                         & 20.00                         & 33.67                         \\ \cmidrule(l){3-13} 
                                  & \multirow{-3}{*}{50\%}        & \emph{KVMerger}                               & \textbf{32.99}                & \textbf{25.31}                & \textbf{18.50}                & \textbf{7.33}                 & \textbf{36.89}                & \textbf{64.00}                & \textbf{26.29}                & \textbf{83.62}                & 20.04                         & \textbf{35.02}                \\ \cmidrule(l){2-13} 
                                  &                               & H2O                                & 30.57                         & 24.48                         & 17.85                         & 7.00                          & 32.17                         & 63.00                         & 25.37                         & 80.89                         & 20.04                         & 33.49                         \\ \cmidrule(l){3-13} 
                                  &                               & CaM                                & 31.06                         & 23.80                         & 18.36                         & 6.00                          & 33.07                         & 62.50                         & 25.23                         & 81.86                         & 18.37                         & 33.36                         \\ \cmidrule(l){3-13}
\multirow{-8}{*}{\rotatebox{90}{\textbf{Llama2-7B-chat}}}  & \multirow{-3}{*}{35\%}        & \emph{KVMerger}                               & \textbf{32.29}                & \textbf{25.24}                & \textbf{19.12}                & \textbf{7.00}                 & \textbf{33.82}                & \textbf{63.50}                & \textbf{25.64}                & \textbf{82.76}                & \textbf{21.09}                & \textbf{34.50}                \\ \midrule 
                                  & \cellcolor[HTML]{DAE8FC}100\% & \cellcolor[HTML]{DAE8FC}Full Cache & \cellcolor[HTML]{DAE8FC}13.21 & \cellcolor[HTML]{DAE8FC}27.59 & \cellcolor[HTML]{DAE8FC}14.42 & \cellcolor[HTML]{DAE8FC}15.25 & \cellcolor[HTML]{DAE8FC}27.44 & \cellcolor[HTML]{DAE8FC}68.50 & \cellcolor[HTML]{DAE8FC}26.69 & \cellcolor[HTML]{DAE8FC}87.42 & \cellcolor[HTML]{DAE8FC}17.15 & \cellcolor[HTML]{DAE8FC}33.07 \\ \cmidrule(l){2-13} 
                                  &                               & H2O                                & 13.39                         & 26.20                         & \textbf{15.01}                         & 15.50                         & 26.40                         & 68.00                         & 25.35                         & 84.73                         & 17.10                         & 32.40                         \\ \cmidrule(l){3-13} 
                                  &                               & CaM                                & 13.30                         & 25.88                         & 13.47                         & 15.00                         & 26.96                         & 67.50                         & 26.06                         & 84.65                         & 16.58                         & 32.16                         \\ \cmidrule(l){3-13} 
                                  & \multirow{-3}{*}{50\%}        & \emph{KVMerger}                               & \textbf{13.46}                & \textbf{26.63}                & 14.4               & \textbf{16.00}                & \textbf{27.29}                & \textbf{68.50}                & \textbf{26.12}                & \textbf{87.48}                & \textbf{17.22}                & \textbf{33.01}                \\ \cmidrule(l){2-13} 
                                  &                               & H2O                                & 12.26                         & 25.52                         & 13.14                         & \textbf{14.50}                & 25.75                         & 67.50                         & 25.59                         & 83.53                         & \textbf{16.35}                & 31.57                         \\ \cmidrule(l){3-13} 
                                  &                               & CaM                                & \textbf{13.43}                         & 25.37                         & 13.58                         & 12.50                          & 25.70                         & 67.50                         & 25.04                         & 84.95                         & 16.34                         & 31.60                         \\ \cmidrule(l){3-13} 
\multirow{-8}{*}{\rotatebox{90}{\textbf{Llama2-13B-chat}}} & \multirow{-3}{*}{35\%}        & \emph{KVMerger}                               & 12.61                & \textbf{26.12}                & \textbf{13.60}                & 14.00                         & \textbf{26.75}                & \textbf{68.00}                & \textbf{26.32}                & \textbf{86.76}                & 16.24                         & \textbf{32.27}                \\ \bottomrule
                                & \cellcolor[HTML]{DAE8FC}100\% & \cellcolor[HTML]{DAE8FC}Full Cache & \cellcolor[HTML]{DAE8FC}31.47 & \cellcolor[HTML]{DAE8FC}26.55 & \cellcolor[HTML]{DAE8FC}21.96 & \cellcolor[HTML]{DAE8FC}25.00 & \cellcolor[HTML]{DAE8FC}39.50 & \cellcolor[HTML]{DAE8FC}61.00 & \cellcolor[HTML]{DAE8FC}26.44 & \cellcolor[HTML]{DAE8FC}83.89 & \cellcolor[HTML]{DAE8FC}30.12 & \cellcolor[HTML]{DAE8FC}38.44 \\ \cmidrule(l){2-13} 
                                  &                               & H2O                                & 29.21                       & 19.91                         & 17.65                         & 8.00                         & 25.50                         & 53.00                         & 19.95                         & 74.55                         & 21.51                         & 29.92                         \\ \cmidrule(l){3-13} 
                                  &                               & CaM                                & 29.57                        & 22.67                         & 19.43                        & 12.00                         & 28.95                         & 58.00                         & 20.17                         & 81.82                         & 21.87                         & 32.72                         \\ \cmidrule(l){3-13} 
                                  & \multirow{-3}{*}{50\%}        & \emph{KVMerger}                               & \textbf{32.44}                & \textbf{24.05}                & \textbf{21.85}               & \textbf{23.00}                & \textbf{31.23}                & \textbf{60.00}                & \textbf{20.87}                & \textbf{84.16}                & \textbf{24.52}                & \textbf{35.79}                \\ \cmidrule(l){2-13} 
                                  &                               & H2O                                & 12.30                         & 5.16                         & 3.64                         & 0.62                & 11.95                        & 37.50                         & 18.99                        & 17.08                         & 14.05               & 13.48                        \\ \cmidrule(l){3-13} 
                                  &                               & CaM                                & 28.77                         & 18.70                         & 17.76                         & 8.50                          & 25.31                         & 45.50                         & 19.72                         & 72.88                         & 17.25                         & 28.27                         \\ \cmidrule(l){3-13} 
\multirow{-8}{*}{\rotatebox{90}{\textbf{Mistral-7B-Instruct}}} & \multirow{-3}{*}{35\%}        & \emph{KVMerger}                               & \textbf{30.77}             & \textbf{20.99}                & \textbf{23.58}                & 23.50                         & \textbf{28.10}                & \textbf{60.5}                & \textbf{19.94}                & \textbf{83.82}                & \textbf{24.13}                         & \textbf{35.04}                \\ \bottomrule
\end{tabular} }
\vspace{-0.5em}
\end{table}

\subsection{Experimental Results on Long-context Tasks}
\textbf{LongBench Results} The evaluation results of nine selected LongBench datasets on Llama2-7B/13B-chat and Mistral-7B-Instruct-v1.0 are shown in Table \ref{tab:LongBench}. We compare the current KV cache compression methods, including H2O, CaM, with our proposed KV merging method \emph{KVMerger} by preserving both $50\%$ and $35\%$ of contexts in the KV cache. Our results demonstrate that \emph{KVMerger} consistently outperforms the other KV cache compression techniques across nearly all selected datasets from LongBench. Notably, the performance gaps between our algorithm and the full KV cache scenario for both Llama2-7B/13B-chat and Mistral-7B-Instruct-v1.0 are significantly smaller than the other KV compression methods. More importantly, our method achieves better evaluation results on several tasks compared to the full cache scenario, highlighting the efficacy and robustness of our approach on long-context tasks. Another interesting finding is that the latest value states merging method, CaM, does not perform well on long-context tasks. This may be attributed to the information loss results from eviction of key states, despite the merging of value states. 

Mistral-7B-Instruct-v1.0 leverages the Grouped-Query-Attention (GQA) technique to optimize KV cache memory usage. In this approach, each key state corresponds to four query states. When applying the H2O method to each key state, rather than duplicating key and value states, we use a single attention map. This attention map is generated by averaging the values of four attention maps formed by the four query states, which determines the states to be evicted. For the \emph{KVMerger} method, we also utilize this singular attention map to select pivotal states, ensuring a fair comparison. Our results indicate a significant performance drop for Mistral-7B-Instruct-v1.0 when using the H2O method. Conversely, \emph{KVMerger} demonstrates the smallest performance decline under both $35\%$ and $50\%$ KV cache budgets, highlighting its efficiency on GQA.
\begin{figure}[t]
    \vspace{-0.5em}
    \centering
    \includegraphics[width=\linewidth]{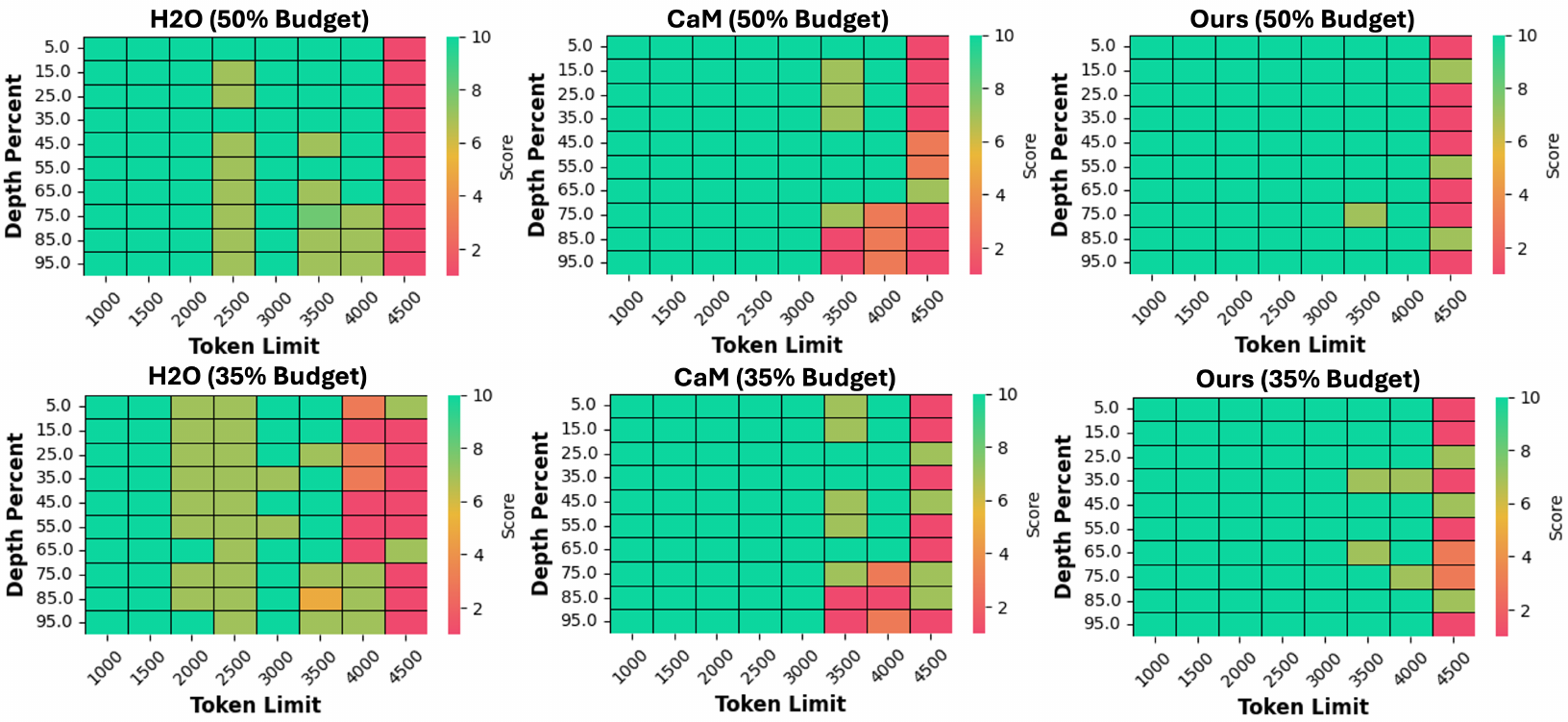}
    \vspace{-1em}
    \caption{The visualization of needle-in-a-haystack test on Llama2-7B-chat with different KV cache compression methods. The x-axis represents the length of contexts, and the y-axis represents the document depth where the needle is inserted. }
    \label{fig:RAG}
    \vspace{-1em}
\end{figure}

\textbf{ZeroScrolls Results} We also evaluate Llama2-7B-chat on ZeroScrolls datasets using different KV cache compression techniques, as shown in Table \ref{tab:ZeroScrolls}. The ZeroScrolls datasets are characterized by an average sample length of approximately 4300 words per topic, closely matching the maximum window size of pre-trained Llama2-7B-chat models. This alignment indicates that the datasets are well-suited for these models, ensuring effective processing and analysis without the risk of truncating important information. Table \ref{tab:ZeroScrolls} demonstrates that our proposed KV cache merging method effectively restores the performance of the Llama2-7B-chat model across all selected ZeroScrolls datasets under both $35\%$ and $50\%$ cache budgets. This suggests that \emph{KVMerger} not only mitigates performance degradation but also optimizes the model's handling of extensive data sequences that approach the model's maximum context window, contributing to more robust outcomes. 

\begin{table}[h]
\vspace{-1em}
\caption{\emph{KVMerger} for Llama2-7B-chat on selected \textbf{ZeroScrolls} datasets} 
\label{tab:ZeroScrolls}
\large
\resizebox{1\textwidth}{!}{
\begin{tabular}{cccccccccc}
\hline
\hline
\textbf{cache budget} & \textbf{Method} & \textbf{gov\_report} & \textbf{SummScreenFD} & \textbf{QMSum} & \textbf{SQuALITY} & \textbf{Qasper} & \textbf{NarrativeQA} & \textbf{BookSumSort} & \textbf{avg.} \\ \hline \hline
\rowcolor[HTML]{DAE8FC} 
100\% & Full Cache & 17.40 & 14.10 & 15.20 & 19.50 & 22.50 & 15.40 & 3.00 & 15.30 \\ \hline
 & H2O & 15.40 & 13.20 & 14.30 & 18.30 & 20.50 & 15.00 & \textbf{3.80} & 14.36 \\ \cline{2-10} 
 & CaM & 15.60 & 13.10 & 13.70 & 18.50 & 20.10 & 15.30 & 3.40 & 14.24 \\ \cline{2-10} 
\multirow{-3}{*}{50\%} &  \emph{KVMerger} & \textbf{17.70} & \textbf{13.80} & \textbf{15.10} & \textbf{19.10} & \textbf{22.50} & \textbf{15.20} & 3.10 & \textbf{15.21} \\ \hline
 & H2O & 14.80 & 11.60 & 14.20 & 17.80 & 17.70 & 14.70 & 3.60 & 13.49 \\ \cline{2-10} 
 & CaM & 15.30 & 11.70 & 13.90 & 18.30 & 17.10 & 14.50 & 3.30 & 13.44 \\ \cline{2-10} 
\multirow{-3}{*}{35\%} &  \emph{KVMerger} & \textbf{16.60} & \textbf{13.80} & \textbf{15.40} & \textbf{18.60} & \textbf{20.40} & \textbf{15.40} & \textbf{3.70} & \textbf{14.84} \\ \hline
\end{tabular}}
\end{table}

\textbf{Needle In A Haystack Results} We also conduct a detailed comparison of \emph{KVMerger} with other KV cache compression techniques on retrieval tasks using the needle-in-a-haystack test. This test involves placing a random fact or statement in the middle of a long context window and assessing the model's ability to retrieve this statement across varying document depths and context lengths to measure performance. Specifically, we test Llama2-7B-chat on document depths ranging from $5\%$ to $95\%$ and context lengths ranging from 1000 to 4500 tokens under both $35\%$ and $50\%$ cache budgets. The corresponding results are illustrated as Figure \ref{fig:RAG}. Our findings indicate that both CaM and our merging algorithm outperform the eviction method H2O. However, our proposed method achieves the highest retrieval performance, consistently delivering high scores across various context lengths and depth percentages. Notably, even when the context length exceeds the pre-trained context length of the Llama2-7B-chat model, our method maintains high scores at specific depth percentages.

\subsection{Ablation Study}
\textbf{Choice of $\sigma$ in Gaussian Kernel Weights} In section 4.2, we set $\sigma$ as the mean of Gaussian kernel weights within each merging set. This definition is based on the empirical observations by testing different values of $\sigma$. Specifically, we apply our proposed KV merging method on Llama2-7B-chat model with various pre-defined $\sigma$ values under $50\%$ cache budget. The experiments results on selected datasets from LongBench are shown as Table \ref{tab:sigma}. We can know that when $\sigma$ equals to 5, the proposed KV cache merging method optimally recovers the model's performance under a $50\%$ cache budget. We then computed $\sigma$ as expressed in Equation 4 for each merging set at different layers and found that the average value of computed $\sigma$ for most layers fluctuates around 5, which aligns with the experiment results in Table \ref{tab:sigma}. 

\begin{table}[h]
\vspace{-1em}
\centering
\small
\caption{\emph{KVMerger} with different $\sigma$ values under $50\%$ cache budget. } 
\label{tab:sigma}
\resizebox{0.75\textwidth}{!}{
\begin{tabular}{ccccccc}
\hline 
\hline
\textbf{$\sigma$} & \textbf{2wikimqa} & \textbf{gov\_report} & \textbf{narrativeqa} & \textbf{pr\_en} & \textbf{multifieldqa\_en} & \textbf{avg.} \\ \hline \hline
\cellcolor[HTML]{FFFFFF}\textbf{0.5} & \cellcolor[HTML]{FFFFFF}29.45 & \cellcolor[HTML]{FFFFFF}24.11 & \cellcolor[HTML]{FFFFFF}18.82 & \cellcolor[HTML]{FFFFFF}6.00 & 35.56 & \cellcolor[HTML]{FFFFFF}22.79 \\ 
\textbf{1} & 31.48 & \textbf{25.52} & \textbf{18.98} & 6.25 & 36.59 & 23.76 \\ 
\textbf{2} & 28.65 & 25.16 & 18.64 & 4.17 & 36.79 & 22.68 \\ 
\textbf{3} & 30.84 & 25.19 & 18.51 & 4.67 & 37.48 & 23.34 \\
\textbf{4} & 31.59 & 25.65 & 18.09 & 5.83 & 36.25 & 23.48 \\ 
\rowcolor[HTML]{DAE8FC} 
\textbf{5} & \textbf{32.99} & 25.31 & 18.50 & \textbf{7.33} & \textbf{36.89} & \textbf{24.20} \\ 
\textbf{6} & 31.69 & 25.39 & 18.45 & 7.83 & 35.82 & 23.84 \\ \hline
\end{tabular}}
\end{table}

\textbf{Choice of Pivotal State in Gaussian Kernel Weighted Merging} As mentioned in Section 4.2, the selection of pivotal state for each merging set is directly related to the performance of \emph{KVMerger}. The inappropriate selection of pivotal states will result in the severe information distortion and even much worse information loss than eviction-based compression algorithms. To show the significance of defining the pivotal state as the state with the biggest aggregated attention scores, we compare it with randomly selecting pivotal state within each merging set by using Llama2-7B-chat model with $50\%$ cache budget. The comparison is shown in Table \ref{tab:pivotal}, from which we can see that randomly selecting pivotal states are detrimental to LLMs' performance on long-context tasks.
\begin{table}[h]
\vspace{-1em}
\centering
\caption{\emph{KVMerger} with different methods of pivotal states selection. } 
\label{tab:pivotal}
\resizebox{0.75\textwidth}{!}{
\begin{tabular}{ccccccc}
\hline 
\hline
\textbf{Pivotal State} & \textbf{2wikimqa} & \textbf{gov\_report} & \textbf{narrativeqa} & \textbf{pr\_en} & \textbf{multifieldqa\_en} & \textbf{avg.} \\ \hline \hline
\rowcolor[HTML]{DAE8FC} 
\textbf{Ours} & 32.99 & 25.31 & 18.50 & 7.33 & 36.89 & 24.20 \\ 
\textbf{Random} & 30.01 & 24.07 & 17.72 & 6.50 & 33.30 & 22.12 \\  \hline
\end{tabular}}
\vspace{-1.5em}
\end{table}
% \vspace{-1em}

\section{Conclusion and Future Work}
In this paper, we propose \emph{KVMerger}, a dynamic KV cache merging method inspired by the observation that key states exhibit high and persistent similarity within each sequence, allowing for layer-wise KV cache compression. We initially abstract the merging set identification problem as a constrained clustering problem and introduce a variant of the AHC algorithm to identify merging sets based on cosine similarities between key states. Furthermore, we implement a Gaussian Kernel weighted merging method to merge key and value states within each merging set. Compared to other KV cache eviction and merging methods, our approach achieves superior results on the LongBench datasets under the same cache budget. Additionally, our method effectively recovers the model's long-context retrieval capabilities, as demonstrated by the needle-in-a-haystack tests.

Future work can explore several avenues to enhance and extend our proposed method. First, investigating the impact of different clustering algorithms and similarity measurements could provide insights into further optimizing the merging sets. Second, applying our method to other LLMs including long-context fine-tuned models and datasets would help assess its generalizability and robustness. Third, exploring hybrid approaches that combine cache merging with other memory management techniques might yield even more efficient solutions for long-context retrieval tasks.

\subsubsection*{Acknowledgments}
This work used Delta system at the National Center for Supercomputing Applications through allocation CIS240055 from the Advanced Cyberinfrastructure Coordination Ecosystem: Services \& Support (ACCESS) program. ACCESS~\citep{access} is an advanced computing and data resource program supported by the U.S. National Science Foundation (NSF) under the Office of Advanced Cyberinfrastructure awards \#2138259, \#2138286, \#2138307, \#2137603 and \#2138296. The Delta advanced computing resource is a joint effort of the University of Illinois Urbana-Champaign and the National Center for Supercomputing Applications, and it is supported by the National Science Foundation (award OAC 2005572) and the State of Illinois. The work also used the Illinois Campus Cluster and NCSA NFI Hydro cluster, which are supported by the University of Illinois Urbana-Champaign and the University of Illinois System.

\bibliography{iclr2024_conference}
\bibliographystyle{iclr2024_conference}

\appendix
\section{Appendix}
\label{appendixA}
\textbf{Lemma 3.1} (Formal version of Lemma 3.1).
\emph{Consider two vectors $\mathbf{k}_{m}$, $\mathbf{k}_{n} \in \mathbb{R}^{1\times d}$. If their cosine similarity is $1$, then the cosine similarity of any $1 \times 2$ vectors, $\mathbf{k}_{m,j} = \left[ k_{m,j}, k_{m,2j+1}\right]^T$ and $\mathbf{k}_{n,j} = \left[ k_{n,2j}, k_{n,2j+1}\right]^T$, formed by the $2j$-th and $(2j+1)$-th elements of $\mathbf{k}_{m}$ and $\mathbf{k}_{n}$, $0 \leq j \leq \frac{d-1}{2}$, is also equal to $1$.}

\emph{Proof.} Since
\begin{equation}
    \textit{similarity}\left( \mathbf{k}_{m}, \mathbf{k}_{n} \right)=1,
\end{equation}
$\mathbf{k}_{m}$ and $\mathbf{k}_{n}$ are collinear. Therefore,
\begin{equation}
    \mathbf{k}_{m} = \alpha \mathbf{k}_{n},
\end{equation}
where $\alpha$ is a scalar. It means 
\begin{align}
    k_{m,2j} & = \alpha k_{n,2j}, \\
    k_{m,2j+1} & = \alpha k_{n,2j+1}.
\end{align}
So,
\begin{equation}
    \left[ k_{m,2j}, k_{m,2j+1} \right]^T = \alpha \left[ k_{n,2j}, k_{n,2j+1} \right]^T.
\end{equation}
As a result,
\begin{equation}
    \textit{similarity}\left( \mathbf{k}_{m,j}, \mathbf{k}_{n,j} \right)=1
\end{equation}

\textbf{Lemma 3.2} (Formal version of Lemma 3.2). 
\emph{Consider integer $j$ such that $0 \leq j \leq \frac{d-1}{2}$. Define the vectors \( \mathbf{k}_{m,j} \) and \( \mathbf{k}_{n,j} \) as \(\mathbf{k}_{m,j} = \left[k_{m,2j},k_{m,2j+1} \right]^T\) and $\mathbf{k}_{n,j} = \left[ k_{n,2j},k_{n,2j+1} \right]^T$, and define the vectors \( \mathbf{k}_{m,j}^{'} \) and \( \mathbf{k}_{n,j}^{'} \) as $\mathbf{k}_{m,j}^{'} = \mathbf{k}_{m,j} / e^{im\theta_{j}}$ and \(\mathbf{k}_{n,j}^{'} = \mathbf{k}_{n,j} / e^{in\theta_{j}} \). If \(  similarity\left( \mathbf{k}_{m,j}, \mathbf{k}_{n,j} \right) = 1\), we have:}
\begin{equation}
    \cos{(m-n)} < \frac{\langle \mathbf{k}_{m,j}^{'}, \mathbf{k}_{n,j}^{'} \rangle}{\|\mathbf{k}_{m,j}^{'}\| \cdot \|\mathbf{k}_{n,j}^{'}\|} \leq 1,
    \nonumber
\end{equation}
where \( \langle \mathbf{k}_{m,j}^{'}, \mathbf{k}_{n,j}^{'} \rangle \) denotes the inner product of \( \mathbf{k}_{m,j}^{'} \) and \( \mathbf{k}_{n,j}^{'} \), and \( \|\mathbf{k}_{m,j}^{'}\| \) and \( \|\mathbf{k}_{n,j}^{'}\| \) denote the norms of \( \mathbf{k}_{m,j}^{'} \) and \( \mathbf{k}_{n,j}^{'} \), respectively.

\emph{Proof.} Since $j$ is an integer obeying $0 \leq j \leq \frac{d-1}{2}$, so
\begin{equation}
    -1 < \frac{-2j}{d} \leq 0.
\end{equation}
And $b$ is set to be $10000$ by default~\cite{su2023roformerenhancedtransformerrotary}. Therefore,
\begin{equation}
    0 < b^{\frac{-2j}{d}} \leq 1,
\end{equation}
which means
\begin{equation}
    \label{eq:9}
    0 < \theta_j \leq 1.
\end{equation}

Now, focus on the similarity between $\mathbf{k}_{m,j}$ and $\mathbf{k}_{n,j}$, and
\begin{equation}
    \frac{\langle \mathbf{k}_{m,j}, \mathbf{k}_{n,j} \rangle}{\|\mathbf{k}_{m,j}\| \cdot \|\mathbf{k}_{n,j}\|}
    =
    \frac{\langle \mathbf{k}_{m,j}^{'}e^{im\theta_{j}}, \mathbf{k}_{n,j}^{'}e^{in\theta_{j}} \rangle}{ \| \mathbf{k}_{m,j}^{'}e^{im\theta_{j}} \| \cdot \| \mathbf{k}_{n,j}^{'}e^{in\theta_{j}} \| }.
    \label{eq:10}
\end{equation}
It is easy to derive that
\begin{align}
    \| \mathbf{k}_{m,j}^{'}e^{im\theta_{j}} \| & = \| \mathbf{k}_{m,j}^{'} \|, \\
    \| \mathbf{k}_{n,j}^{'}e^{in\theta_{j}} \| & = \| \mathbf{k}_{n,j}^{'} \|,
\end{align}
since the exponential terms do not change the vectors' magnitude.

Then, substitute the complex forms of $\mathbf{k}_{m,j}^{'}$ and $\mathbf{k}_{n,j}^{'}$, $\mathbf{k}_{m,j}^{'} = k_{m,2j}^{'} + i k_{m,2j+1}^{'}$ and $\mathbf{k}_{n,j}^{'} = k_{n,2j}^{'} + i k_{n,2j+1}^{'}$, respectively, back into $\langle \mathbf{k}_{m,j}^{'}e^{im\theta_{j}}, \mathbf{k}_{n,j}^{'}e^{in\theta_{j}} \rangle$ and obtain
\begin{align}
    \langle \mathbf{k}_{m,j}^{'}e^{im\theta_{j}}, \mathbf{k}_{n,j}^{'}e^{in\theta_{j}} \rangle
    = \langle \left( k_{m,2j}^{'} + i k_{m,2j+1}^{'} \right)e^{im\theta_j}, \left( k_{n,2j}^{'} + i k_{n,2j+1}^{'} \right)e^{in\theta_j} \rangle.
    \label{eq:13}
\end{align}
From Euler equation, Equation~\ref{eq:13} can be further expanded as
\begin{align}
    & \langle \mathbf{k}_{m,j}^{'}e^{im\theta_{j}}, \mathbf{k}_{n,j}^{'}e^{in\theta_{j}} \rangle \nonumber \\
    = & k_{m,2j}^{'}k_{n,2j}^{'}\cos{(m\theta_j)}\cos{(n\theta_j)} + k_{m,2j+1}^{'}k_{n,2j+1}^{'}\sin{(m\theta_j)}\sin{(n\theta_j)} \nonumber \\
      & - k_{m,2j}^{'}k_{n,2j+1}^{'}\cos{(m\theta_j)}\sin{(n\theta_j)} - k_{m,2j+1}^{'}k_{n,2j}^{'}\sin{(m\theta_j)}\cos{(n\theta_j)} \nonumber \\
      & + k_{m,2j}^{'}k_{n,2j}^{'}\sin{(m\theta_j)}\sin{(n\theta_j)} + k_{m,2j+1}^{'}k_{n,2j+1}^{'}\cos{(m\theta_j)}\cos{(n\theta_j)} \nonumber \\
      & + k_{m,2j}^{'}k_{n,2j+1}^{'}\sin{(m\theta_j)}\cos{(n\theta_j)} + k_{m,2j+1}^{'}k_{n,2j}^{'}\cos{(m\theta_j)}\sin{(n\theta_j)} \nonumber \\
    = & k_{m,2j}^{'}k_{n,2j}^{'}\cos{\left[ (m-n)\theta_j \right]} + k_{m,2j+1}^{'}k_{n,2j+1}^{'}\cos{\left[ (m-n)\theta_j \right]} \nonumber \\
      & + k_{m,2j}^{'}k_{n,2j+1}^{'}\sin{\left[ (m-n)\theta_j \right]} + k_{m,2j+1}^{'}k_{n,2j}^{'}\sin{\left[ (m-n)\theta_j \right]}.
    \label{eq:14}
\end{align}
Substitute Equation~\ref{eq:14} back into Equation~\ref{eq:10}, and
\begin{align}
    \frac{\langle \mathbf{k}_{m,j}, \mathbf{k}_{n,j} \rangle}{\|\mathbf{k}_{m,j}\| \cdot \|\mathbf{k}_{n,j}\|}
    = &
    \frac{k_{m,2j}^{'}k_{n,2j}^{'} + k_{m,2j+1}^{'}k_{n,2j+1}^{'}}{\|\mathbf{k}_{m,j}^{'}\| \cdot \|\mathbf{k}_{n,j}^{'}\|}\cos{\left[ (m-n)\theta_j \right]} \nonumber \\
    & + \frac{k_{m,2j}^{'}k_{n,2j+1}^{'} - k_{m,2j+1}^{'}k_{n,2j}^{'}}{\|\mathbf{k}_{m,j}^{'}\| \cdot \|\mathbf{k}_{n,j}^{'}\|}\sin{\left[ (m-n)\theta_j \right]} \nonumber \\
    = & 
    \frac{\mathbf{k}_{m,j}^{'} \cdot \mathbf{k}_{n,j}^{'}}{\|\mathbf{k}_{m,j}^{'}\| \cdot \|\mathbf{k}_{n,j}^{'}\|}\cos{\left[ (m-n)\theta_j \right]} 
    + \frac{\mathbf{k}_{m,j}^{'} \times \mathbf{k}_{n,j}^{'}}{\|\mathbf{k}_{m,j}^{'}\| \cdot \|\mathbf{k}_{n,j}^{'}\|}\sin{\left[ (m-n)\theta_j \right]}.
    \label{eq:15}
\end{align}
Let $\phi$ be angle between $\mathbf{k}_{m,j}^{'}$ and $\mathbf{k}_{n,j}^{'}$, then Equation~\ref{eq:15} can be rewrite as
\begin{align}
    \frac{\langle \mathbf{k}_{m,j}, \mathbf{k}_{n,j} \rangle}{\|\mathbf{k}_{m,j}\| \cdot \|\mathbf{k}_{n,j}\|} \nonumber
    = & \cos{\phi}\cos{\left[ (m-n)\theta_j \right]} + \sin{\phi}\sin{\left[ (m-n)\theta_j \right]} \nonumber \\
    = & \cos{\left[ \phi - (m-n)\theta_j \right]}.
\end{align}
Since the similarity between $\mathbf{k}_{m,j}$ and $\mathbf{k}_{n,j}$ nearly equals 1 as we assumed, it can be obtained that
\begin{equation}
    \phi = \left( m-n\right)\theta_j.
\end{equation}
From Equation~\ref{eq:9},
\begin{align}
    0 < &\phi \leq m-n, \quad \text{if } m>n, \\
    m-n \leq &\phi < 0, \qquad \quad \  \text{if } m<n.
\end{align}
As a result,
\begin{equation}
    \cos{(m-n)} < \frac{\langle \mathbf{k}_{m,j}^{'}, \mathbf{k}_{n,j}^{'} \rangle}{\|\mathbf{k}_{m,j}^{'}\| \cdot \|\mathbf{k}_{n,j}^{'}\|} \leq 1.
\end{equation}

\end{document}

%% file: math_commands.tex
\usepackage{amsmath,amsfonts,bm}

\def\eqref#1{equation~\ref{#1}}
\def\1{\bm{1}}

\DeclareMathAlphabet{\mathsfit}{\encodingdefault}{\sfdefault}{m}{sl}
\SetMathAlphabet{\mathsfit}{bold}{\encodingdefault}{\sfdefault}{bx}{n}